\documentclass[letterpaper, 10 pt, conference]{IEEEtran}
\IEEEoverridecommandlockouts
\usepackage[T1]{fontenc}
\usepackage{textcomp}
\usepackage[bookmarks=true]{hyperref}
\usepackage{amsmath}
\usepackage[capitalise]{cleveref}
\usepackage{siunitx}
\usepackage[numbers, sort]{natbib}
\usepackage{amsmath,amssymb,amsfonts}
\usepackage{algorithmic}
\usepackage{graphicx}
\usepackage{enumerate}
\usepackage{textcomp}
\usepackage{xcolor}
\usepackage{cuted}
\usepackage[compatibility=false]{caption}
\usepackage{subcaption}
\usepackage{glossaries}
\usepackage{booktabs}
\usepackage{hyperref}
\usepackage{pifont}
\usepackage{adjustbox} 
\usepackage{bm}
\def\BibTeX{{\rm B\kern-.05em{\sc i\kern-.025em b}\kern-.08em
    T\kern-.1667em\lower.7ex\hbox{E}\kern-.125emX}}
\usepackage{eso-pic}

\newcommand{\cmark}{\ding{51}}
\newcommand{\xmark}{\ding{55}} 
\newacronym{mpc}{MPC}{Model Predictive Controller}
\newacronym{ads}{ADS}{Autonomous Driving Systems}
\newacronym{cpu}{CPU}{Central Processing Unit}
\newacronym{lidar}{LiDAR}{Light Detection and Ranging}
\newacronym{ftg}{FTG}{Follow-The-Gap}
\newacronym{fsd}{FSD}{Formula Student Driverless}
\newacronym{iac}{IAC}{Indy Autonomous Challenge}
\newacronym{sota}{SotA}{State of the Art}
\newacronym{map}{MAP}{Model- and Acceleration-based Pursuit}
\newacronym{dtr}{DTR}{Delaunay Triangulation-based Racing}
\newacronym{imu}{IMU}{Inertial Measurement Unit}
\newacronym{ccma}{CCMA}{Curvature Corrected Moving Average}
\newacronym{mctr}{MCTR}{Midpoint-Corrected Triangulation for Racing}
\newacronym{pp}{PP}{Pure Pursuit}

\begin{document}

\title{MCTR: Midpoint-Corrected Triangulation for Autonomous Racing via Digital Twin Simulation in CARLA}

\author{
Junhao Ye$^{1*}$,
Cheng Hu$^{1, 2*}$,
Yiqin Wang$^{1*}$,
Weizhan Huang$^{1}$,
Nicolas Baumann$^{3}$,
Jie He$^{4}$,\\
Meixun Qu$^{4}$,
Lei Xie$^{1\dagger}$,
and Hongye Su$^{1}$%
\thanks{$^*$ \textbf{Equal contribution}. }%
\thanks{$^{1}$ Department of Control Science and Engineering, Zhejiang University, Hangzhou, China.}%
\thanks{$^{2}$ Ningbo Innovation Center, Zhejiang University, Ningbo, China.}%
\thanks{$^{3}$ Center for Project-Based Learning, D-ITET, ETH Zurich.}%
\thanks{$^{4}$ Technische Universität Wien, Vienna, Austria.}%
\thanks{$^\dagger$ Corresponding authors: {\tt\small  lxie@iipc.zju.edu.cn.}}%
}
\maketitle

\begin{strip} 
\vspace{-2.5cm}
\centering
\includegraphics[angle=0,origin=c,width=\textwidth, height=6.8cm]{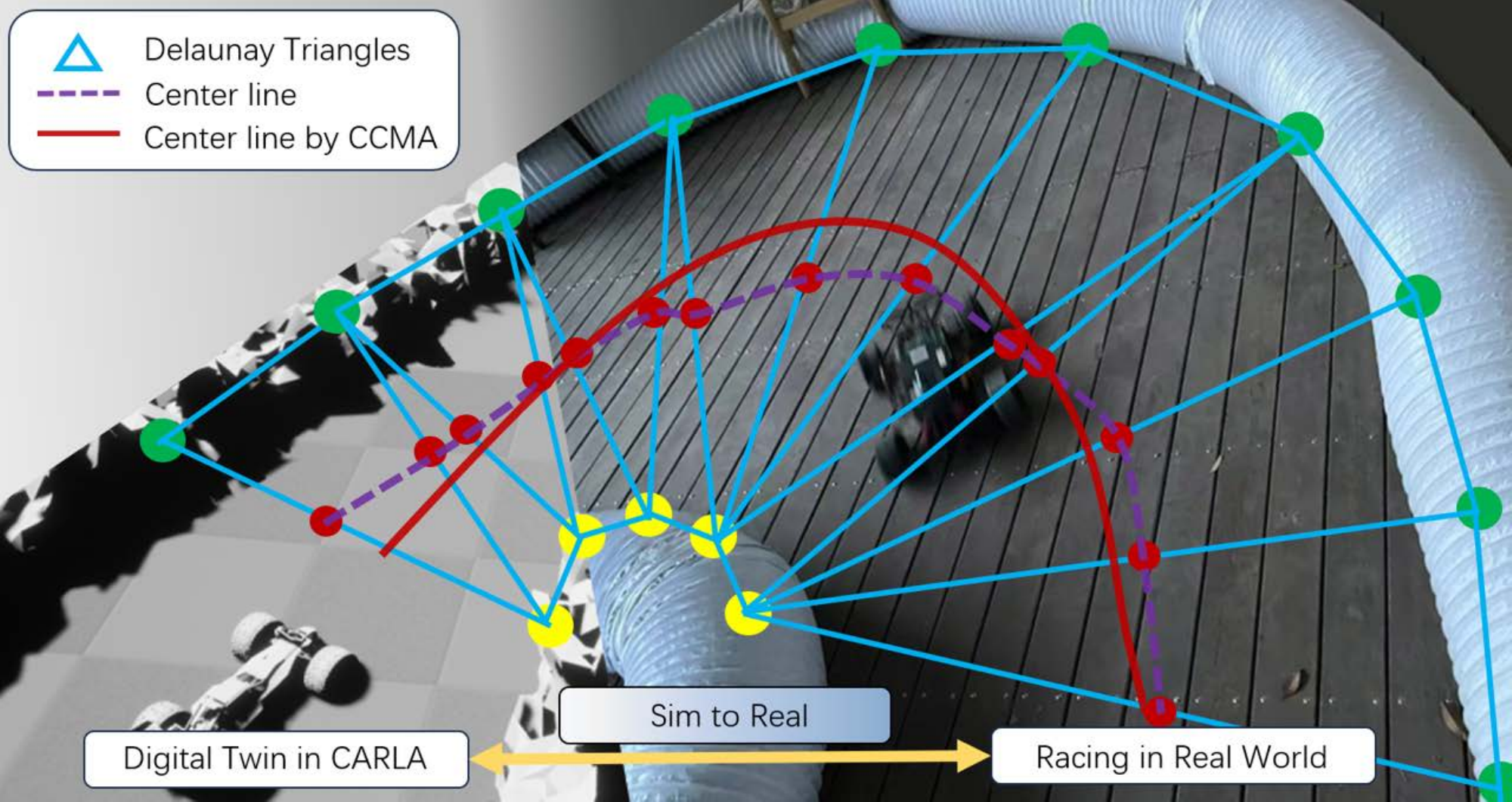}
\vspace{-0.25cm}
\captionof{figure}{ Illustration of the proposed \gls{mctr} controller. Lidar-scanned track data is triangulated, and midpoints of opposing edges are extracted to form the initial centerline. Trajectory stability and smoothness are enhanced through \gls{ccma} smoothing. The left image shows the algorithm being tested in Carla simulation, while the right presents its real-world deployment, demonstrating effective sim-to-real transfer.}
\label{fig:graphical_abstract}
\vspace{-0.25cm}
\end{strip}
\glsresetall

\begin{abstract}
In autonomous racing, reactive controllers eliminate the computational burden of the full \emph{See-Think-Act} autonomy stack by directly mapping sensor inputs to control actions. This bypasses the need for explicit localization and trajectory planning. A widely adopted baseline in this category is the \gls{ftg} method, which performs trajectory planning using LiDAR data. Building on \gls{ftg}, the \gls{dtr} algorithm introduces further enhancements. However, \gls{dtr}’s use of circumcircles for trajectory generation often results in insufficiently smooth paths, ultimately degrading performance. Additionally, the commonly used F1TENTH-simulator for autonomous racing competitions lacks support for 3D \gls{lidar} perception, limiting its effectiveness in realistic testing. To address these challenges, this work proposes the \gls{mctr} algorithm. \gls{mctr} improves trajectory smoothness through the use of \gls{ccma} and implements a digital twin system within the CARLA simulator to validate the algorithm’s robustness under 3D \gls{lidar} perception. The proposed algorithm has been thoroughly validated through both simulation and real-world vehicle experiments.

\end{abstract}

\glsresetall

\section{Introduction}
Most high-performance autonomous driving systems, including those deployed in F1TENTH~\cite{okelly2020f1tenth}, \gls{iac}~\cite{indyautonomous}, and \gls{fsd} \cite{amz_fullstack}, follow the conventional \textit{See-Think-Act} pipeline~\cite{siegwart_amr}. In this architecture, the system relies on accurate global localization to determine its position relative to a pre-mapped track, and executes control actions to follow an offline-planned trajectory. This approach enables near-optimal racing performance under ideal conditions by leveraging precise vehicle dynamics models and detailed prior knowledge~\cite{betz2022autonomous}. However, it fundamentally depends on two critical assumptions: (1) the availability of a globally optimal reference path, and (2) the ability to maintain centimeter-level localization accuracy at all times~\cite{forzaeth}.

Worse still, these assumptions become fragile under high-speed and high-side-slip conditions of autonomous racing~\cite{dikici2025learning}. Operating near the physical limits of the vehicle magnifies the effects of model mismatch and control latency~\cite{hu2025sparse}. Even brief localization dropouts or minor deviations from the planned trajectory can lead to drastic performance degradation or loss of control. These challenges expose the vulnerability of planning- and localization-centric autonomy stacks, highlighting the need for more robust frameworks that can adapt in real time without relying entirely on global priors~\cite{tognoni2025dtr}.

In contrast to localization-dependent pipelines, reactive controllers such as \gls{ftg} \cite{ftg} operate without requiring global maps or full-state estimation. These approaches infer control commands directly from real-time local sensor data, enabling rapid response to dynamic conditions. \gls{dtr}~\cite{tognoni2025dtr} is one such method that extracts candidate waypoints by geometrically processing triangulated track boundaries from \gls{lidar} scans. While \gls{dtr} achieves impressive cornering performance by leveraging limited global context, its reliance on centerline extraction introduces vulnerability. In particular, the resulting trajectories can lack smoothness and are sensitive to variations in the \gls{lidar} scanning plane, making them susceptible to failure in the presence of sensor misalignments or noise.

These limitations are further exacerbated by the fact that many racing simulators, including the F1TENTH Simulator, only support 2D \gls{lidar} and lack the ability to simulate high-resolution 3D point clouds~\cite{o2020f1tenth}. As a result, algorithms that rely on richer perception—such as those intended for full-scale vehicles equipped with 3D \gls{lidar}—cannot be effectively developed or validated in these environments. This limitation reduces the representational fidelity of simulation and hinders the seamless transition from virtual to physical testing. These shortcomings reinforce the need for a digital-twin simulation framework that accurately mirrors real-world sensing and dynamics, supports modular sensor configurations, and scales across vehicle platforms. Such a capability is essential for enabling the next generation of robust, perception-driven autonomous racing systems~\cite{r-carla}.

To address these challenges, we propose the \gls{mctr} reactive controller. Built upon \gls{dtr}, \gls{mctr} introduces a midpoint-based opposing-edge selection method to extract a smoother and more stable centerline, which is further refined using the \gls{ccma} filter. Our algorithm is capable of filtering 3D LiDAR point cloud data and reprojecting it into 2D LiDAR-like data based on angle and distance, thereby ensuring compatibility with both 2D and 3D LiDAR sensors. The proposed approach is evaluated on a physical 1:10 scale platform and within an enhanced CARLA-based digital-twin simulator that supports both 2D \gls{lidar} range scans and full 3D \gls{lidar} point cloud inputs. This dual-sensor compatibility enables the construction of a comprehensive Sim-to-Real validation pipeline, demonstrating scalability from small-scale to full-scale platforms. The key contributions of this work are as follows:
\begin{enumerate}[I] 
\item \textbf{Centerline Extraction with Curvature Smoothing}: Building upon \gls{dtr}, the proposed \gls{mctr} algorithm introduces an opposing-edge midpoint extraction strategy based on driving direction and incorporates curvature smoothing via the \gls{ccma} algorithm. This leads to smoother, curvature-preserving centerlines with improved stability and robustness across diverse track conditions. The generated centerlines are then used by \gls{pp} to compute control commands for the vehicle.

\item \textbf{Sim-to-Real Validation with Multi-Sensor Support}: A CARLA-based digital-twin framework is developed and integrated, enabling simulation of both 2D and 3D \gls{lidar} data with an emphasis on ensuring behavioral consistency between simulation and real-world operation, thereby facilitating reliable sim-to-real transfer.

\item \textbf{Open-Source Implementation}: All algorithms, simulation tools, and experimental configurations are released to facilitate reproducibility and further research.The code is available
at: \href{https://github.com/ZJU-DDRX/MCTR.}{github.com/ZJU-DDRX/MCTR.}
\end{enumerate}

\section{Related Work}
Autonomous racing has emerged as a prominent benchmark for evaluating high-speed control and perception capabilities in autonomous systems. While model-based approaches such as \gls{mpc} and end-to-end learning methods have been extensively studied, they typically require global maps, accurate system models, or large training datasets—requirements often impractical in real-time racing scenarios. Therefore, a core focus in this domain is the development of reactive controllers, which generate control commands directly from local sensor inputs, without relying on pre-defined maps. This makes them particularly effective in highly dynamic scenarios like autonomous racing, where global localization may be unreliable or temporarily lost.

However, evaluating such controllers directly on real vehicles under high-speed conditions poses significant safety risks and logistical challenges. In these situations, pushing control algorithms to their limits can result in vehicle damage or unsafe behaviors.

To address this, digital twin simulators have become essential tools for safely and realistically validating control strategies before real-world deployment. Among them, CARLA stands out due to its support for both 2D and 3D \gls{lidar} data, and high-fidelity modeling of vehicle dynamics. CARLA provides a scalable and flexible environment for testing reactive controllers under diverse conditions, enabling the development of robust and transferable control strategies, and helping to bridge the gap between simulation and real-world performance.

\subsection{\gls{sota} controllers in Autonomous Racing}
\subsubsection{Model-based Methods in Autonomous Racing}
Model-based approaches, such as \gls{mpc} \cite{williams2017information,hewing2020cautious} and \gls{map} \cite{becker2022model}, are commonly used in autonomous racing to optimise trajectories. These methods rely on accurate models of the vehicle and environment, and require global map knowledge. While model-based methods can deliver near-optimal performance under ideal conditions, they are often computationally intensive and sensitive to modelling inaccuracies, making them impractical for real-time control in dynamic and partially observable environments.

\subsubsection{End-to-End Learning Approaches}
End-to-end learning methods, such as those proposed by Kendall et al. \cite{kendall2019learning} and Betz et al. \cite{betz2022survey}, aim to directly map raw sensor data to control outputs. While these methods have shown promising results in simulated environments, they typically require large amounts of labelled training data and significant computational resources. Moreover, they tend to be sensitive to distribution shifts and sensor noise, which can lead to unreliable or unstable behaviour in highly dynamic and safety-critical scenarios like autonomous racing. Additionally, the lack of transparency and interpretability compared to traditional geometric or model-based approaches further limits their applicability in real-time systems.

In contrast to both model-based and end-to-end learning approaches, the proposed \gls{mctr} framework offers a practical and efficient alternative tailored for autonomous racing. Unlike MPC, \gls{mctr} does not rely on global map knowledge or precise vehicle models, which significantly reduces computational complexity and enhances robustness to modeling inaccuracies. Compared to end-to-end methods, \gls{mctr} leverages geometric reasoning and reactive planning, ensuring greater interpretability and stability under sensor noise and high-speed conditions. 

\subsection{Reactive Control Approaches}
Reactive control plays a central role in autonomous racing, where environments are highly dynamic and often lack reliable global maps. These methods directly convert raw sensor data into control commands, making them well-suited for real-time, high-speed operation.

One widely adopted reactive control method is the \gls{ftg} approach ~\cite{ftg}, which maps local sensor measurements to steering and velocity commands by identifying safe gaps in the environment. While \gls{ftg} is simple and effective, it can suffer in complex track geometries or narrow racing scenarios due to its purely reactive nature.

A more structured method is \gls{dtr}~\cite{tognoni2025dtr}, which uses delaunay triangulation on \gls{lidar} scans to extract centerlines. While more reliable than \gls{ftg}, \gls{dtr} relies on a predefined geometric division of the track into left, right, and centre regions. This increases sensitivity to varying track layouts and often requires manual parameter tuning for different environments. In addition to these structural limitations, \gls{dtr} also faces challenges during the path smoothing stage. It also relies on B-splines~\cite{deBoor1978} smoothing, which may introduce oscillations in sharp or noisy regions. 

To address these issues, \gls{mctr} uses opposing edge detection for centerline extraction without segmentation and applies \gls{ccma} smoothing to enhance stability under challenging conditions.

\subsection{Simulation Platforms in Autonomous Racing}
The F1TENTH simulator \cite{okelly2020f1tenth} has become a benchmark for autonomous racing research, especially for small-scale vehicles using 2D \gls{lidar} sensors. However, its scalability is limited for full-size racecars, and it cannot simulate dense 3D \gls{lidar} point clouds, restricting its real-world applicability. In contrast, CARLA offers superior scalability for full-scale racecars, supporting customizable models and sensors, including dense 3D \gls{lidar} point clouds, significantly improving its ability to replicate real-world conditions~\cite{del2021autonomous}. Built on Unreal Engine, CARLA enables high-fidelity simulations of vehicle dynamics and sensor behaviours across diverse environments. Its flexibility supports various sensor configurations like \gls{lidar}, cameras, and IMUs, crucial for testing autonomous systems in complex, real-world-like racing scenarios. Additionally, CARLA’s digital-twin compatibility allows for accurate simulation environments using real-world data, further bridging the gap between simulation and real-world testing~\cite{r-carla}.

\subsection{Summary and Comparison}
Compared to existing methods, \gls{mctr} offers several key advantages. It combines the strengths of \gls{dtr}'s local sensor-based centerline extraction with \gls{ccma}'s effective path smoothing, resulting in more stable and smoother trajectories. Additionally, our method's adaptability to both 2D and 3D \gls{lidar} sensors, along with its successful validation in a digital twin environment (CARLA), distinguishes it from other approaches, contributing to its robustness and scalability. A table comparing our method with related work is presented in \Cref{tab:rw}.

\begin{table} [h]
\centering
    \begin{adjustbox}{max width=\columnwidth}
    \begin{tabular}{l|c|c|c|c}
        \toprule
          & \textbf{Mapless} & \textbf{Sensor Setup} & \textbf{Simulator} & \textbf{CPU usage}\\
        \midrule
               \acrshort{map} \cite{becker2023map}  & \xmark
        & \textit{Full-Stack} & F1TENTH & Heavy \\
        \acrshort{mpc} \cite{williams2017information, hewing2020cautious} & \xmark & \textit{Full-Stack} & F1TENTH & Heavy \\
        \midrule
         \acrshort{ftg} \cite{ftg, forzaeth}  & \cmark & 2D \gls{lidar} & F1TENTH & Light \\
        \acrshort{dtr} \cite{tognoni2025dtr} & \cmark & 2D \gls{lidar} & F1TENTH &Light \\
        \acrshort{mctr} \textbf{(Ours)}  &
        \cmark & 2D/3D \gls{lidar} & F1TENTH/CARLA & Light \\
        \bottomrule
    \end{tabular}
    \end{adjustbox}
\caption{Comparison of racing algorithms. \textbf{Mapless}  means no need for a predefined map. Due to the need for localization and state estimation in model-based control methods, CPU usage can be very high.}
\label{tab:rw}
\end{table}
\section{Methodology}
The system we proposed comprises two key components: (A) centerline extraction with a control method, and (B) Sim-to-Real validation with Multi-Sensor support. This integrated framework enables mapless autonomous racing based on 2D/3D \gls{lidar} perception data and achieves seamless sim-to-real transfer.

Our method is entirely based on \gls{lidar} perception data, extracting the track centerline by detecting the midpoints between the opposite track boundaries. The extracted centerline is then used for trajectory tracking via \gls{pp}, with steering angles and velocities computed using a physical vehicle model, which is similar to the control strategy in \gls{dtr}; hence, we choose \gls{dtr} as our baseline.

Besides, our algorithm is compatible with both 2D and 3D \gls{lidar} point cloud data. For sensor deployment, we utilise the \emph{HOKUYO UST-10LX} for 2D \gls{lidar} and the \emph{Livox Mid-360} for 3D \gls{lidar} data collection. Before real-world experiments, it is essential to validate the algorithm's performance in simulation environments. For a 2D \gls{lidar} simulation, we employ the well-established F1TENTH simulator. However, as the F1TENTH simulator lacks compatibility with 3D point cloud data, we adopt the digital-twin approach proposed in the $\mathcal{R}$-CARLA framework to reconstruct the track environment. Furthermore, we introduce a lightweight conversion that projects 3D point clouds into 2D \gls{lidar} format, enabling the algorithm to ingest diverse sensor modalities.
\subsection{Centerline Extraction With Control Method}
\subsubsection{\textbf{{Centerline Extraction}}}
Similar to the approach of Kabzan et al. \cite{amz_fullstack}, we use Delaunay triangulation to extract the centerline from range measurements of the track boundaries. To do this, we perform secondary sampling of the 2D \gls{lidar} scans according to the boxed method from \cite{stahl2019ros}, followed by Delaunay triangulation of these \gls{lidar} points. Distinguishing from the \gls{dtr} approach, we add detection for opposing edges to extract midpoints of the selected edges as the centerline. Furthermore, the three geometric divisions mentioned in \gls{dtr} are deemed unnecessary in our algorithm, avoiding the need for cumbersome parameter adjustments for different track conditions.

Consequently, our algorithm enhances the credibility of the original path before trajectory smoothing through the following methods:
\begin{itemize}
    \item \textbf{Opposing Edge}: If a triangle edge has two vertices located on opposite sides of the driving direction, it is identified as an opposing edge.
    \item \textbf{Faraway Point}: If the distance between the current centerpoint and the next point exceeds a specified threshold, it is considered a faraway point.
    \item \textbf{Pseudo Point}: If a centerpoint lies at the track’s end, it is treated as an unqualified pseudo point.
\end{itemize}

\begin{figure}[htb]
    \centering
    \includegraphics[width=\linewidth,  trim=0cm 0cm 0cm 0cm, clip]{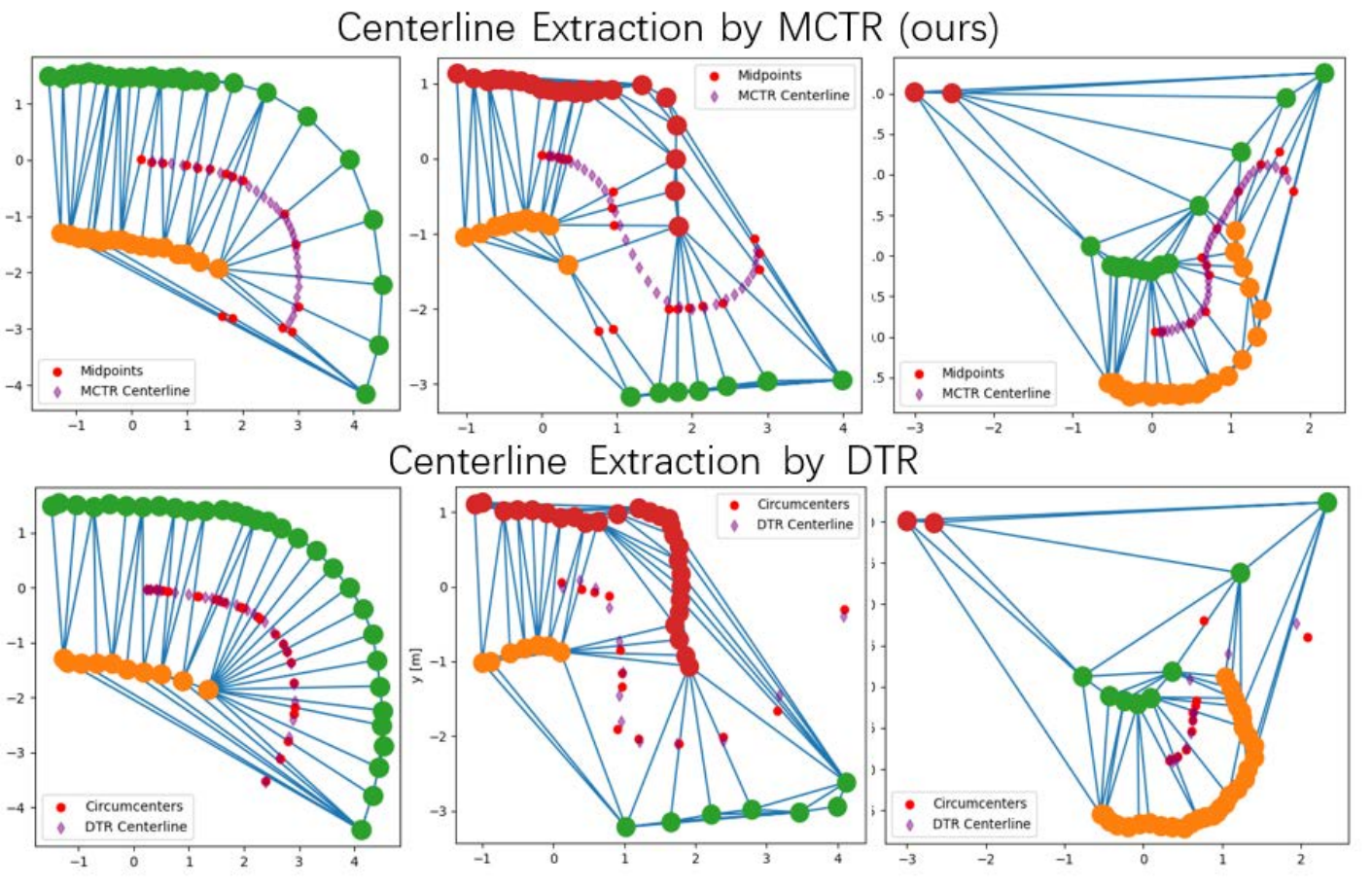}
    \caption{Our \gls{mctr} algorithm is shown above in the image, while the \gls{dtr} algorithm is shown below. The image demonstrates that our algorithm has stronger robustness due to the absence of geometric parameter constraints.}
    \label{fig:about midline}
\end{figure}

Only opposing edges are selected to extract the centerpoints. After performing linear interpolation for the faraway points and eliminating the pseudo points, our algorithm achieves high robustness with a set of parameters that work well across various track conditions. A fig showcasing the robustness of our algorithm in three corners compared to \gls{dtr} is provided in \Cref{fig:about midline}.

Once the centerpoints are identified, the next step is to connect them to form an approximate centerline. The centerline extraction process follows the method outlined in \gls{dtr}, retaining only the centerpoints that lie ahead of the vehicle and within the \gls{lidar} scan range. A greedy nearest-neighbor algorithm is then employed to sequentially construct the centerline.

Before smoothing the ordered centerline points using the Savitzky-Golay filter, we first smooth the curvature using the \gls{ccma} algorithm. The use of \gls{ccma} ensures that curvature fluctuations between adjacent points are minimized, which effectively reduces vehicle oscillations and significantly increases the achievable speed along the trajectory. Compared to traditional denoising or moving average methods, \gls{ccma} directly operates on curvature and enforces curvature continuity, making it particularly suitable for local path planning scenarios where small-scale geometric smoothness critically impacts control stability. By penalizing both first- and second-order discrete derivatives, \gls{ccma} generates smoother and dynamically feasible trajectories while maintaining proximity to the original path. The optimization problem is as follows:
\begin{align}
E = & \sum_{i=1}^{n} \left\| q_i - p_i \right\|^2 \nonumber \\
  & + \lambda \sum_{i=2}^{n-1} \left\| q_{i+1} - 2q_i + q_{i-1} \right\|^2 \\
  & + \mu \sum_{i=3}^{n-2} \left\| q_{i+2} - 4q_{i+1} + 6q_i - 4q_{i-1} + q_{i-2} \right\|^2 \nonumber
\end{align}
\noindent
where the first term preserves fidelity to the original points $p_i$, the second term promotes first derivative (velocity) continuity, and the third term enforces second derivative (curvature) smoothness. The weighting parameters $\lambda$ and $\mu$ control the trade-off between trajectory smoothness and tracking accuracy. This curvature-aware smoothing formulation improves tracking performance and ride comfort, particularly in tight or cluttered environments.

\subsubsection{\textbf{Control Method}}
Once the centerline is obtained, the controller calculates the control commands—steering angle and target velocity—to guide the race car along the track. Conventional methods such as \gls{ftg} rigidly steer toward the largest gap at a fixed lookahead, often producing aggressive oscillations in the resulting trajectory.

To generate smoother motion, we adopt a pure-pursuit framework, where a lookahead distance is determined, and an appropriate point on the centerline is selected as the tracking point. The lookahead distance balances stability and responsiveness; a larger distance is chosen for high-speed driving (e.g., straights) to enhance stability, while shorter distances are used for low-speed driving (e.g., corners). The lookahead distance is computed as follows:

\begin{equation}
L_d = k_v v + L_{\min}
\end{equation}
where $k_v$ is the speed-dependent gain coefficient and $L_{\min}$ is the minimum safety distance.

The steering angle is computed using the kinematic bicycle model. The model assumes a single-track approximation of the race car, and the steering command is derived geometrically by calculating the angular deviation between the vehicle’s current heading and the direction of the lookahead point. This angle is then transformed into a steering angle using the bicycle model’s formulation, similar to \cite{pure_pursuit}.

\begin{equation}
\delta = \arctan\!\left(\frac{2\,L\,\sin\alpha}{L_d}\right)
\end{equation}
\noindent
where $L$ is the vehicle wheelbase, $\alpha$ is the heading error (i.e., the angle between the current orientation and the lookahead point vector), and $L_d$ is the lookahead distance.

Velocity commands are computed based on the vehicle’s dynamic constraints, ensuring safe navigation through curves by considering track curvature and available tire-road grip. First, an estimate of the friction coefficient \(\mu\) is obtained by pulling the car laterally along the center of mass with a spring scale. The maximum admissible speed \(v_{adm}\) is derived from the balance between lateral acceleration—directly influenced by the local curvature—and the frictional limits of the vehicle, as inspired by the first approximation used for velocity generation in \cite{heilmeier2020mincurv}:

\begin{equation}
v_{target} = \sqrt{\mu a_{max}^{y} \kappa^{-1}}.
\label{speed}
\end{equation}

Due to the smoother curvature generated by the \gls{mctr} algorithm, the vehicle can maintain higher speeds through curves, resulting in an increased maximum attainable speed.

\subsection{Sim-to-Real Validation with Multi-Sensor Support}
\subsubsection{\textbf{Digital Twin Environment}}
To build the digital twin environment, the \emph{Livox Mid-360} was initially used to extract track information and generate point cloud data from the real world. The point clouds were then processed using a spherical outlier rejection algorithm described in the $\mathcal{R}$-CARLA framework, followed by Poisson disk sampling \cite{ulichney1988dithering} to simplify the point cloud. The simplified point cloud is used to generate a mesh via the Ball Pivoting algorithm \cite{r-carla}. The mesh is deployed into the UE4 editor to create the digital twin map for 3D algorithm simulation.

\begin{figure}[htb]
    \centering
    \includegraphics[width=\linewidth, height=6.5cm]{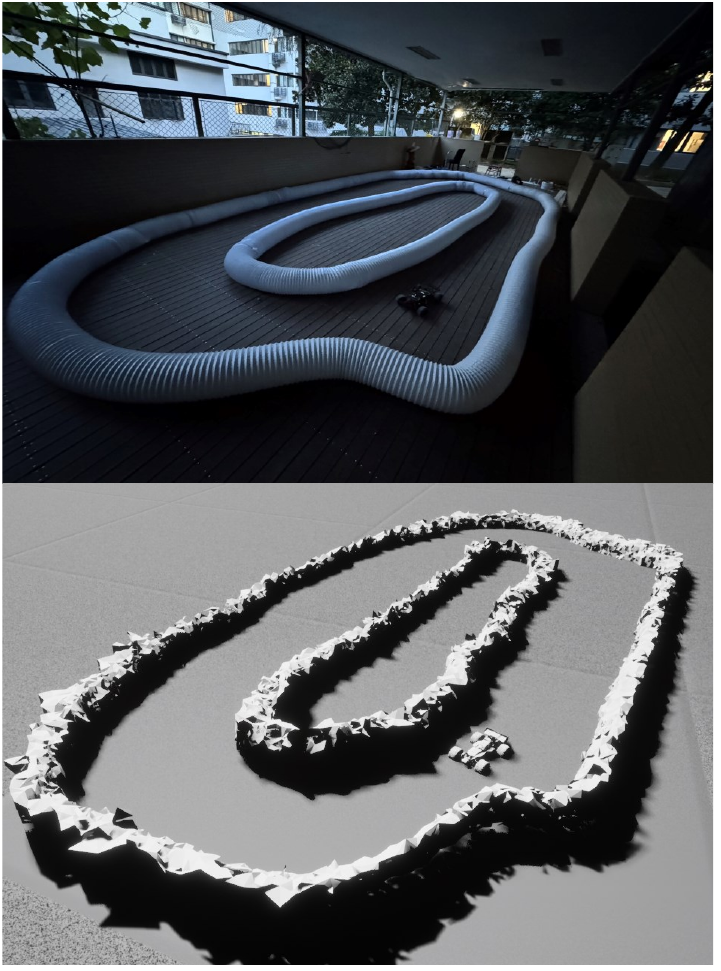}
    \caption{The above image is a real-world RGB image, while the image below demonstrates the digital twin of the 1:1 scale track scene and car model, enabling the car equipped with a 3D \gls{lidar} to be validated in the simulation environment.}
    \label{fig:Sim2real}
\end{figure}

To model the real race car, we scale the existing models to the size of the F1TENTH car and modify the relevant mechanical constraints to ensure that the simulation closely matches the physical model. The digital twin image based on \cite{r-carla} is shown in \Cref{fig:Sim2real}.

\begin{figure}[htb]
    \centering
    \includegraphics[width=\linewidth, trim=0cm 0cm 0cm 0cm, clip]{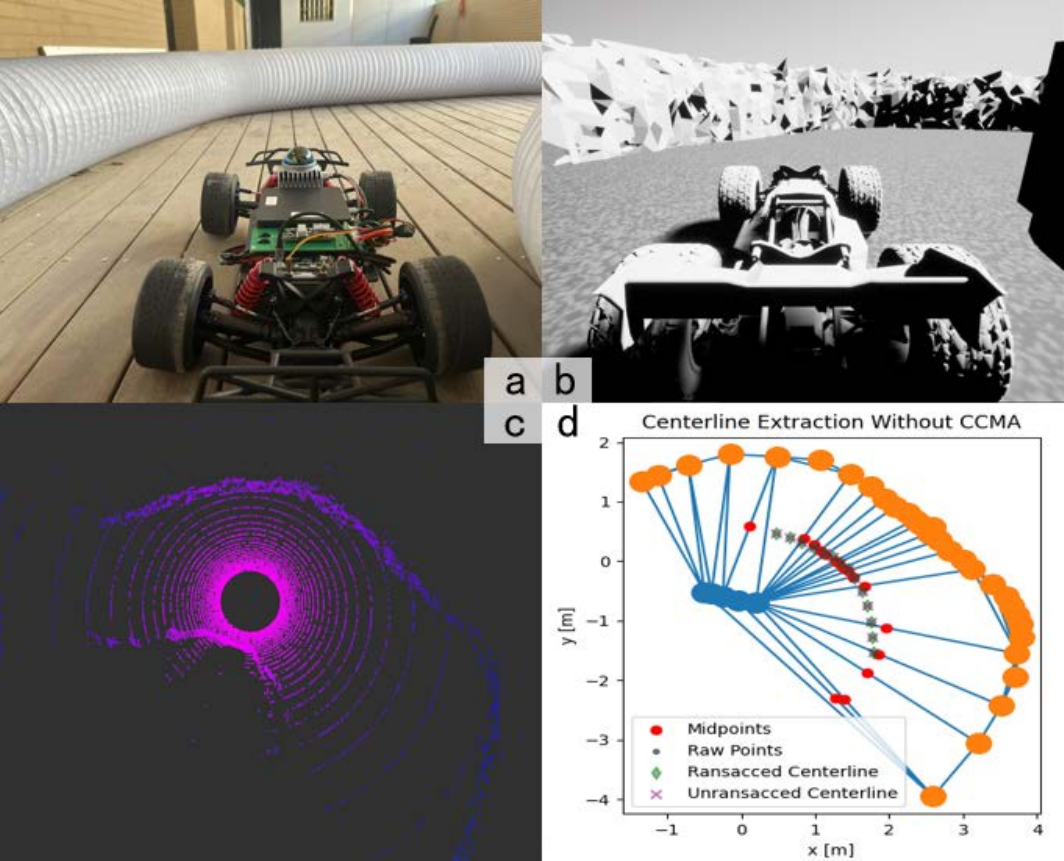}
    \caption{Figure (a) shows the physical race track, figure (b) shows the car's viewpoint in the CARLA simulator, figure (c) is a 3D \gls{lidar} point cloud, and figure (d) is the route map generated by \gls{mctr} based on the perception data.}
    \label{fig:3D lidar}
\end{figure}

Initially deployed in the CARLA simulator, the algorithm was adapted to process 3D \gls{lidar} point cloud data by handling 3D \gls{lidar} data with independent \gls{imu}s. Only track-relevant point clouds are projected, selecting the appropriate z-axis point cloud to construct pseudo-2D \gls{lidar} data for algorithm input. This method prevents the vehicle body posture (e.g., tilt) from affecting point extraction during racing. The simulation of the \gls{mctr} algorithm in a 3D scene is shown in the \Cref{fig:3D lidar}.

In the 3D simulation, communication with the CARLA server is done through the \texttt{carla-ros-bridge}, which controls the throttle and brake commands. Due to the imprecision of the velocity controller provided by \texttt{carla-ros-bridge}, this work introduces feedback control for both throttle and brake after obtaining the velocity control commands. The formulas for the throttle and brake control outputs are given as follows:
\begin{equation}
\begin{bmatrix}
\text{Throttle}(t) \\
\text{Brake}(t)
\end{bmatrix}
=
\begin{bmatrix}
K_p^{t} & K_i^{t} & K_d^{t} \\
K_p^{b} & K_i^{b} & K_d^{b}
\end{bmatrix}
\cdot
\begin{bmatrix}
e(t) \\
\sum_{i=0}^{t} e(i) \\
\frac{e(t) - e(t-1)}{\Delta t}
\end{bmatrix}
\end{equation}
 where $K_p^i$, $K_i^i$, and $K_d^i$ (i=t/b) represent the proportional, integral, and derivative gains for the throttle/brake PID controller, respectively. The control error at time $t$, denoted as $e(t)$, is defined as the difference between the target velocity $v_{\text{target}}$ and the actual velocity $v_{\text{actual}}$.
\subsubsection{\textbf{Validation with Multi-Sensor}}
Our method is grounded on range measurements first segmented by distance and azimuth from 2D \gls{lidar}. We validate these measurements in the well-established F1TENTH simulator and confirm their effectiveness on a physical platform equipped with a \emph{HOKUYO UST-10LX}, achieving highly satisfactory results.

\begin{figure}[htb]
    \centering
    \includegraphics[width=\linewidth, height=4cm , trim=0cm 0cm 0cm 0cm, clip]{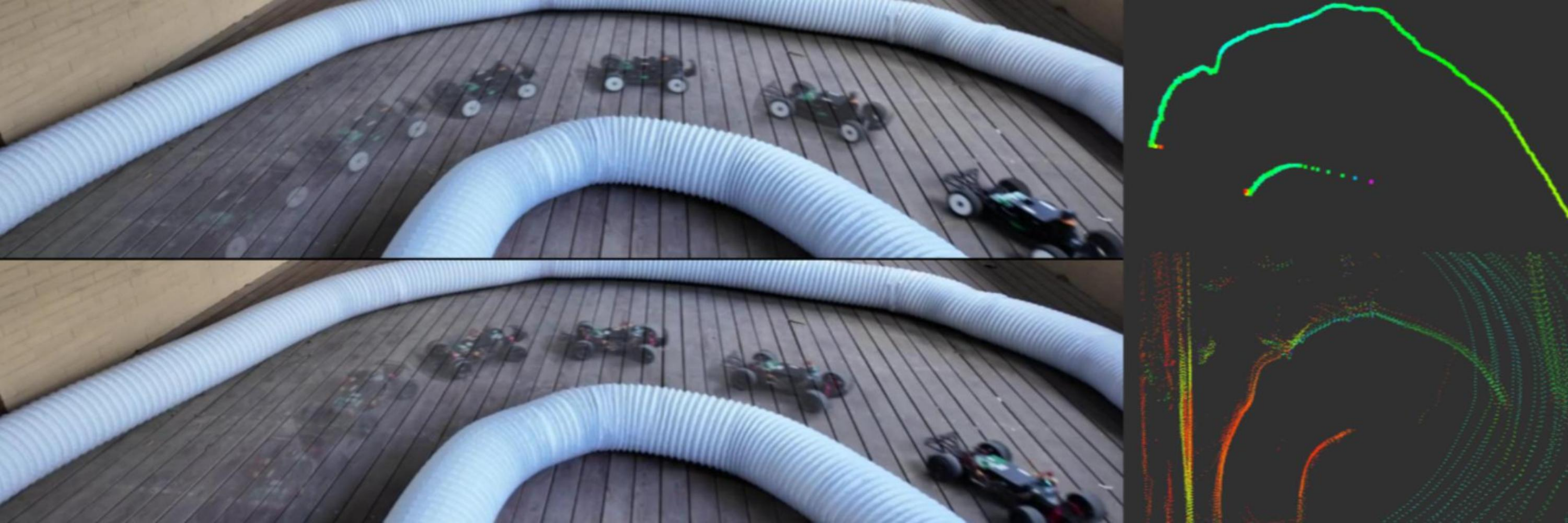}
    \caption{Figure contrasts the physical poses and corresponding scans of 2D and 3D \gls{lidar} traversing the same curve, which demonstrates  nearly identical performance.}
    \label{fig:2D_3D lidar}
\end{figure}

To extend the same pipeline to 3D \gls{lidar}, we introduce a lightweight conversion. Firstly, as described earlier, we exploit a CARLA-based digital twin to acquire 3D point clouds in simulation. We then filter points by Z-height, retaining only those within a prescribed vertical band. Each remaining point is projected onto the XY-plane; its Euclidean distance to the origin and its azimuth are computed, yielding a pseudo-2D \gls{lidar} scan in (range, angle) format compatible with MCTR. Finally, we deploy a \emph{Livox Mid-360} on the vehicle and demonstrate that the proposed adaptation fully satisfies the algorithmic validation requirements. The real-world setups for both Multi-Sensor experiments are shown in \Cref{fig:2D_3D lidar}

\section{Result}
To comprehensively evaluate the proposed algorithm, we conduct both simulation and real-world experiments. The simulation results, including detailed ablation studies, provide insights into the contributions of individual components such as the \gls{ccma} smoothing strategy and the midpoint-based extraction method. These studies highlight their respective roles in enhancing the overall system performance. In parallel, real-world experiments are designed to assess robustness across different reactive controllers and sensing configurations. Notably, the comparable performance observed between 2D and 3D \gls{lidar} setups in physical deployments underscores the practical viability and generalizability of the digital twin-based development workflow.
\subsection{Comparison in Simulator}
\begin{figure}[htb]
    \centering
    \includegraphics[width=\linewidth, trim=0cm 0cm 0cm 0cm, clip]{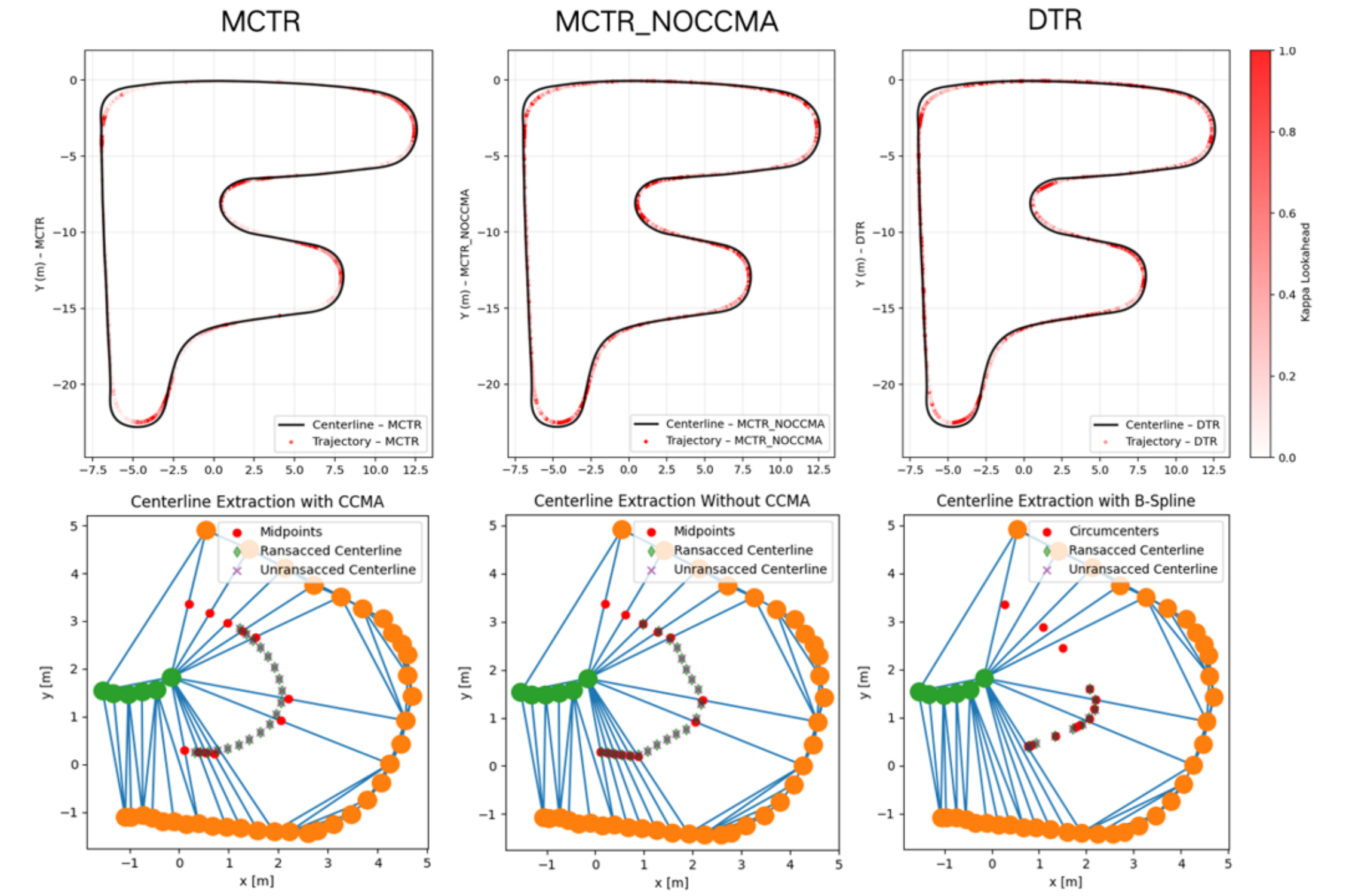}
    \caption{From left to right, the figure shows a simulation comparison of the \gls{mctr}, \gls{mctr} (no \gls{ccma}), and \gls{dtr} algorithms on the \emph{F}-shaped track. The top part reflects the variation of the lookahead point curvature with the car's motion, while the bottom image displays the generation of trajectory points when navigating through sharp corners on the track.}
    \label{fig:F_test}
\end{figure}

\Cref{fig:F_test} illustrates simulation results on the F-shaped track within the F1TENTH simulator, comparing the performance of the complete \gls{mctr} method, its ablated variant without the \gls{ccma} module (denoted as "Ours no \gls{ccma}"), and the baseline \gls{dtr} method. All three algorithms successfully complete the course; however, subtle differences in curvature smoothness during cornering are apparent, as reflected in the varying intensities of the lookahead-point curvature color map in \Cref{fig:F_test}. To highlight these differences more clearly, we also present the extracted centerlines in cornering scenarios. The results show that the \gls{mctr} algorithm produces smoother curvature transitions and demonstrates notable advantages in both the accuracy and density of centerline extraction. 

Moreover, to quantitatively assess the differences among the three algorithms, we evaluate three specific metrics, the lap time \textbf{$t_{lap}$}, the average curvature error \textbf{$\varepsilon_k$},  and the average centerline lateral error jerk \textbf{$J_{lat}$}.

Under F-shaped track conditions, the proposed \gls{mctr} method delivers substantial performance improvements over the baseline \gls{dtr} controller. Specifically, it reduces the lap time \textbf{$t_{lap}$} from 28.73 s to 23.39 s. In addition, \gls{mctr} achieves a 20.00\% reduction in the curvature error \textbf{$\varepsilon_k$} and a 24.83\% reduction in the lateral jerk cost \textbf{$J_{lat}$}, compared to \gls{dtr}. These improvements indicate enhanced path smoothness, which effectively suppresses oscillatory behavior and contributes to better overall driving stability.

Furthermore, we performed an ablation study on the \gls{mctr} algorithm by setting up a variant without the \gls{ccma} smoothing module. The experimental results show that the \gls{mctr} algorithm without \gls{ccma} still outperforms \gls{dtr} in \textbf{$t_{lap}$}, which corresponds to a 17.9\% improvement. However, compared to \gls{dtr}, the smoothness and stability of the \gls{mctr} trajectory without \gls{ccma} slightly decrease, although the difference is minimal, likely due to the challenges posed by higher speeds. The detailed data are presented in the \Cref{tab:performance_comparison}

\begin{table}[h]
\centering
\begin{adjustbox}{max width=\columnwidth} % Resize table to fit within the page width
\begin{tabular}{l|c|c|c}
    \toprule
     & \textbf{Ours} & \textbf{Ours (no CCMA)} & \textbf{\acrshort{dtr}} \\ \midrule
    {\textbf{$t_{lap}$}} [s] $\downarrow$ & $\boldsymbol{23.39 \pm 0.60}$&  $23.59\pm 0.15$ & $28.73 \pm 0.38$ \\ 
    {\textbf{$\varepsilon_k$}} [$\text{m}^{-1}$] $\downarrow$ & $\boldsymbol{0.012 \pm 0.002}$ & $0.018 \pm 0.003$ & $0.015 \pm 0.002$ \\ 
    {\textbf{$J_{lat}$}} [$\text{m/s}^3$] $\downarrow$ &$\boldsymbol{0.115 \pm 0.02}$ & $0.142 \pm 0.03$ & $0.153 \pm 0.03$ \\ 
    \bottomrule
\end{tabular}
\end{adjustbox}
\caption{Performance comparison of \textit{Reactive} controllers: Ours (\gls{mctr}), Ours (\gls{mctr} without CCMA), and \gls{dtr}. Metrics are reported in the $\mu \pm \sigma$ format ($\mu$: mean, $\sigma$: standard deviation). Bold values indicate the best performance for each metric. Key performance indicators include lap time ($t_{lap}$), curvature error ($\varepsilon_k$), and centerline lateral error jerk ($J_{lat}$).}
\label{tab:performance_comparison}
\end{table}

Overall, the full \gls{mctr} outperforms both its variant without \gls{ccma} and the baseline \gls{dtr} across all evaluated metrics, achieving better lap time, smoother trajectories, and improved stability. These advantages are also visually evident in the cornering behavior shown in \Cref{fig:F_test}.

\subsection{Comparison in Real World}
Real-world experiments were conducted on three relevant algorithms: \gls{mctr}, \gls{dtr}, and \gls{ftg}. Their performance was compared across various racing scenarios, including the O-shaped track shown in \Cref{fig:Sim2real}. As discussed in the velocity calculation method \eqref{speed}, velocity is inversely proportional to the curvature of the predicted points, meaning that smoother trajectories should lead to shorter lap times \textbf{$t_{lap}$}. Therefore, we configured four distinct scenarios and recorded the \textbf{$t_{lap}$} for each algorithm on the real-world track we set up to validate the smoothness of the algorithms. Additionally, we recorded CPU usage and passability metrics to further assess the algorithms' performance. The specific results are shown in \Cref{tab:performance_comparison in different maps}.

In terms of lap times, \gls{mctr} demonstrates a relative advantage over \gls{dtr} and \gls{ftg}, even in a simple track test. If the track difficulty is further increased, such as by setting track boundaries with more pronounced curvature variations, it is anticipated that \gls{mctr} will perform even better. \gls{mctr} also exhibits good passability; however, the improvement in passability comes at the cost of higher CPU usage, as \gls{mctr} uses \gls{ccma} for smoothing, which increases the algorithm's overhead. Compared to the passability metric, the cost of CPU usage for \gls{mctr} is considered acceptable.
\begin{table}[ht]
\centering
\begin{adjustbox}{max width=\columnwidth} % Resize table to fit within the page width
\begin{tabular}{l|c|c|c}
    \toprule
     & \textbf{\gls{mctr}} & \textbf{DTR} & \textbf{FTG} \\ \midrule
    {\textbf{$t_{O}$}} [s] $\downarrow$ & $\boldsymbol{7.39}$ &  $7.82$ & $8.13$ \\ 
    {\textbf{$t_{F}$}} [s] $\downarrow$ & $\boldsymbol{9.77}$ &  $9.88$ &$12.91$  \\ 
    {\textbf{$t_{M}$}} [s] $\downarrow$ & \textbf{$9.59$} &  $\boldsymbol{9.28}$ & $N/A$  \\ 
    {\textbf{$t_{W}$}} [s] $\downarrow$ & $\boldsymbol{10.18}$ &  $N/A$ & $N/A$ \\ 
    {\textbf{$CPU$}} [$\%$] $\downarrow$ & \textbf{$57.22$} & $25.65$ & $\boldsymbol{12.46}$  \\ 
    {\textbf{$PASS$}} [$\%$] $\uparrow$ & $\boldsymbol{95}$ & $80$ & $45$  \\ 
    \bottomrule
\end{tabular}
\end{adjustbox}
\caption{Table records the performance of different algorithms across various tracks. The difficulty increases sequentially, with average lap time and CPU usage tested on the O, F, M, and W-shaped tracks. Each track was tested for five laps, and the \emph{PASS} data indicates the percentage of uninterrupted laps (out of a total of 20) in which no minor collisions occurred, demonstrating the robustness of the algorithms}
\label{tab:performance_comparison in different maps}
\end{table}

Furthermore, we also tested the performance of the \gls{mctr} algorithm using a 3D \gls{lidar} sensor. The \Cref{fig:2D_3D lidar} demonstrated 2D \gls{lidar} and 3D \gls{lidar} running in the physical world, achieving approximate results, which is due to the efficiency improvement brought by our digital twin simulator. The vehicle equipped with the 3D \gls{lidar} sensor achieves a lap time \textbf{$t_{lap}$} of just 7.90 seconds in Carla simulation, closely matching the 7.23 seconds recorded in the real-world experiment. This strong consistency verifies the effectiveness of the digital twin approach. However, the transition from 2D to 3D \gls{lidar} increased CPU utilisation from 57.22 \% to 60.71 \%, as the point cloud data from the 3D \gls{lidar} is denser, consuming more CPU resources.

\section{Conclusion}
This paper addresses critical challenges in reactive control for autonomous racing by introducing the \gls{mctr} algorithm. Building upon prior work in \gls{dtr}, our approach fundamentally enhances trajectory smoothness and robustness through two key innovations: (1) a midpoint-based opposing-edge selection strategy for stable centerline extraction, and (2) curvature-constrained moving average (\gls{ccma}) filtering for generating physically consistent racing lines. Unlike conventional approaches, \gls{mctr} eliminates the need for geometric track segmentation while maintaining compatibility with both 2D and 3D \gls{lidar} sensors through our novel projection methodology.

Comprehensive validation demonstrates \gls{mctr}'s significant performance advantages over state-of-the-art reactive controllers. In simulation, \gls{mctr} achieved a lap time of 23.39 seconds on the F-shaped track—a 19\% improvement over \gls{dtr}—while reducing curvature error by 20\% and lateral jerk by 25\%. Real-world testing on a 1:10 scale platform confirmed these gains, with \gls{mctr} consistently outperforming baselines across diverse track configurations and achieving a 95\% passability rate in complex scenarios. The developed CARLA-based digital twin framework proved essential for robust sim-to-real transfer, particularly for 3D \gls{lidar} validation, where our projection method maintained performance parity with dedicated 2D sensing.

Despite its strengths, \gls{mctr} currently depends on structured tracks with clear boundaries. Future work will explore stereo vision for unstructured environments and improve the digital twin with more accurate vehicle dynamics.

\bibliographystyle{IEEEtranN}
\bibliography{ref}

\end{document}